\pdfoutput=1
\documentclass[11pt]{article}

\usepackage[final]{acl} 

\usepackage{times}
\usepackage{latexsym}
\usepackage[T1]{fontenc}
\usepackage[utf8]{inputenc}
\usepackage{microtype}
\usepackage{inconsolata}
\usepackage{graphicx}

\usepackage{booktabs}
\usepackage{multirow}
\usepackage{subcaption}
\usepackage{placeins}
\usepackage{amsmath}

\newcommand{\stitle}[1]{\vspace{1ex} \noindent{\bf #1.}}

\usepackage{xspace}

\newcommand{\datasetname}{\textsc{FoodPuzzle}\xspace}

\title{ \mbox{\includegraphics[scale=0.035]{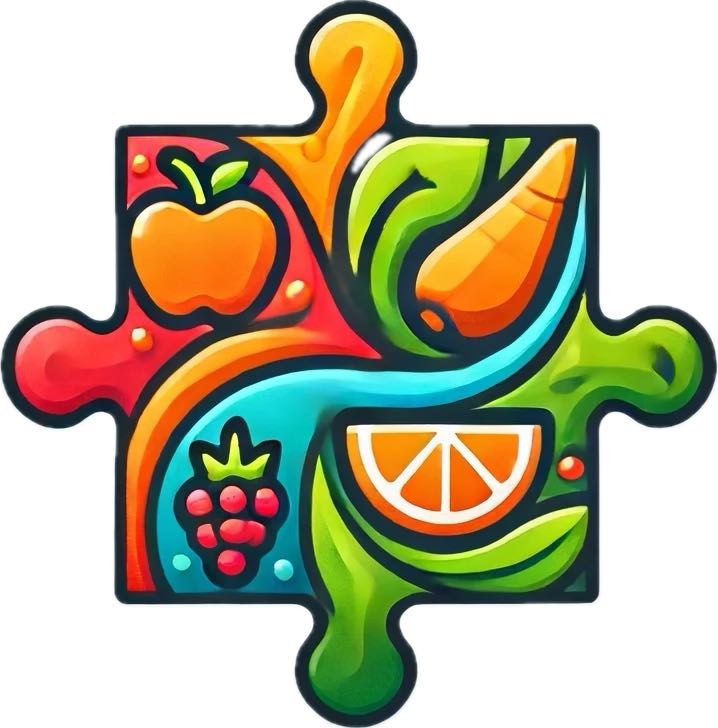}\datasetname : Developing Large Language Model Agents as Flavor Scientists}}
\author{
    \textbf{Tenghao Huang\textsuperscript{1}}, \textbf{Donghee Lee\textsuperscript{2*}}, \textbf{John Sweeney}\thanks{Equal contribution.}, \textbf{Jiatong Shi}, 
    \textbf{Emily Steliotes\textsuperscript{2}}, \\ \textbf{Matthew Lange\textsuperscript{2}}, \textbf{Jonathan May\textsuperscript{1}}, \textbf{Muhao Chen\textsuperscript{2}} \\[1ex]
    \textsuperscript{1}University of Southern California \\
    \textsuperscript{2}University of California, Davis \\
    \texttt{tenghaoh@usc.edu}
}
\begin{document}
\maketitle
\begin{abstract}
Flavor development in the food industry is increasingly challenged by the need for rapid innovation and precise flavor profile creation. Traditional flavor research methods typically rely on iterative, subjective testing, which lacks the efficiency and scalability required for modern demands. 
This paper presents three contributions to address the challenges. Firstly, we define a new problem domain for scientific agents in flavor science, conceptualized as the generation of hypotheses for flavor profile sourcing and understanding. 
To facilitate research in this area, we introduce the \datasetname, a challenging benchmark consisting of 978 food items and 1,766 flavor molecules profiles. We propose a novel
Scientific Agent approach, integrating in-context learning and retrieval augmented techniques to generate grounded hypotheses in the domain of food science. Experimental results indicate that our model significantly surpasses traditional methods in flavor profile prediction tasks, demonstrating its potential to transform flavor development practices.
\end{abstract}

\section{Introduction}


The burgeoning demand for novel and appealing flavors in the food industry necessitates expedited innovation cycles without compromising on quality \cite{Hofmann2018-kz}.
Traditionally, the process of ingredient sourcing and evaluation for flavor development has been reliant on manual assessment and human expertise. In addition to being labor-intensive, and subjective, this process also  relies on time-consuming iterations, hindering the pace of flavor development \cite{Engeseth2018-bl, Hofmann2018-kz, Patel2019-bp}. Additionally,  complex flavor interactions, together with the vast array of available ingredients, present formidable challenges for food scientists in optimizing flavor profiles \cite{Hamilton2020-by}. 

\begin{figure}[t!]
    \includegraphics[width=0.4\textwidth]{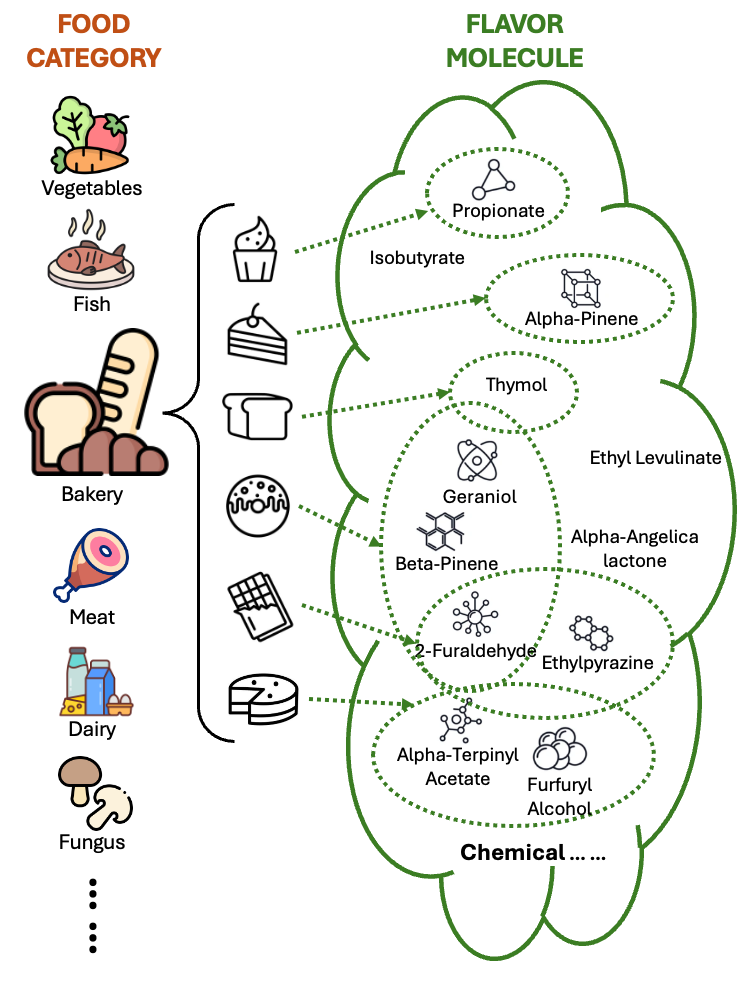}
    \small
    \centering
    \caption{Flavor is determined by diverse flavor molecules. Sourcing and identifying these molecules from various foods is a time-consuming and resource-intensive task for food scientists. Understanding these connections is crucial in flavor science for developing and enhancing food products to ensure appealing taste experiences for consumers. In this work, we explore how LLMs can assist in this process. }
    \label{fig:data_hierarchy}
\end{figure}

\begin{figure*}[t]
    \small
    \centering
    \includegraphics[width=0.9\textwidth]{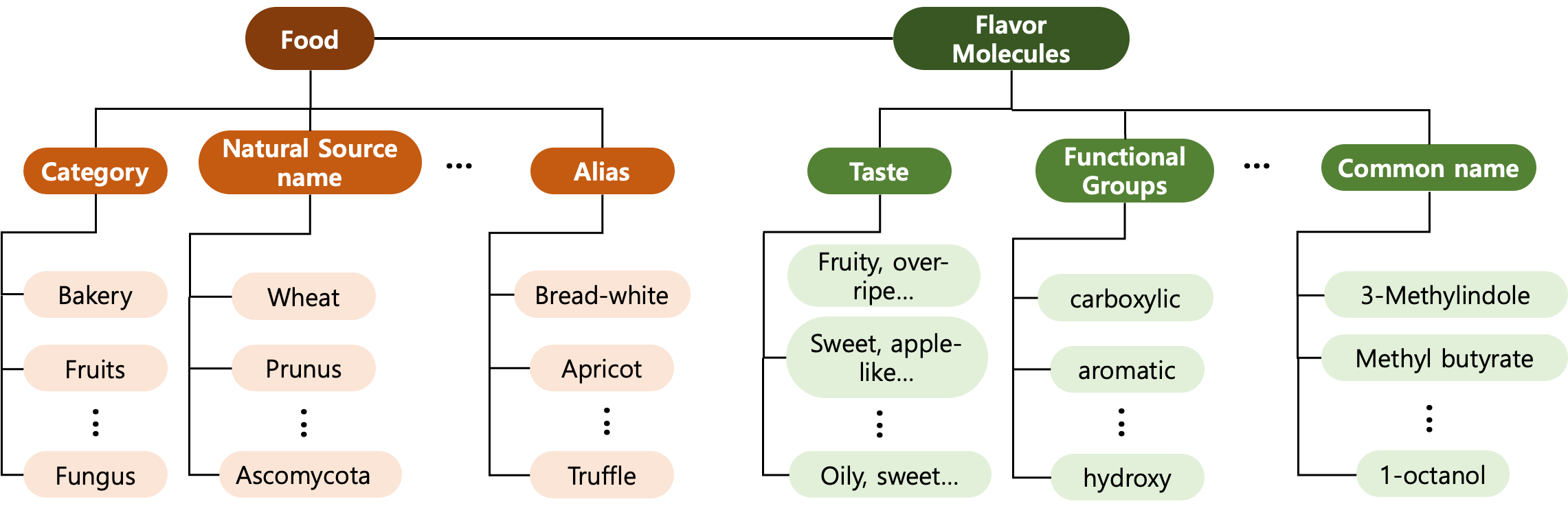}
    \caption{A high level overview of the \datasetname data hierarchy}
    \label{fig:categorization}
\end{figure*}

Recent advancements in Large Language Models (LLMs) have revolutionized traditional scientific methodologies across a broad spectrum of disciplines. In fields ranging from biology and chemistry to physics and material sciences, researchers have employed LLM technologies to analyze complex datasets, uncover hidden patterns, and 
narrow down the search spaces of scientific problems \cite{horawalavithana-etal-2022-foundation,Singhal2022LargeLM, o2023bioplanner, song-etal-2023-matsci, Hong2024DataIA}. This trend highlights the transformative potential of LLMs in reshaping traditional paradigms and catalyzing breakthroughs. 
This paper explores how the integration of LLMs can streamline and enhance the process of evaluating ingredient sources for their flavor-giving potential. However, the application of LLMs in ingredient sourcing and flavor formulation is hindered by a significant barrier: the absence of high-quality, domain-specific datasets. Most LLMs are trained on generic data from the Web, which often lacks the detailed, specialized knowledge required for accurate flavor profile understanding and prediction.
In this paper, we delineate three primary contributions that collectively advance the application of LLMs in the flavor science. Our \textit{first} contribution is to formulate flavor profile sourcing and understanding as a LLM agent problem. Flavor sourcing and understanding are inherently complex tasks, involving an open-ended exploration of vast ingredient arrays to pinpoint components that deliver desired profiles. Scientific agents excel at navigating these complexities by leveraging their capacity to identify relevant evidence and reason within large context spaces effectively. When integrated with external knowledge sources, these agents can perform the labor-intensive tasks of flavor sourcing and understanding with enhanced efficiency and precision.



Our \textit{second} contribution is the creation of the \includegraphics[scale=0.02]{figs/icon.jpg} \datasetname  dataset. To gather data for our study on flavor profile prediction and understanding, we turn to FlavorDB \cite{garg2018flavordb}. FlavorDB is a comprehensive database that includes detailed information on 25,595 flavor molecules, of which 1,766 are specifically noted for their presence in 978 natural ingredients. Figure~\ref{fig:categorization} presents an overview of the structure of our constructed dataset. We propose two tasks that mirror real-world scientific study. The first task is Molecular Food Prediction (MFP), which challenges the model to predict possible food items based on a given set of flavor molecules. The second task is completing the molecular profile (MPC) which requires the identification of flavor molecules likely present in a specified food item. 


As we realize the challenge of existing approaches for solving the FoodPuzzle task,
our \textit{third} contribution is the development of 
a comprehensive scientific agent method for proposing grounded hypotheses of flavor profile prediction tasks\footnote{Our codes and the dataset will be publicly released upon acceptance.}. 
By integrating in-context learning (ICL) and the Retrieval-Augmented Generation (RAG) approach, our method enhances the model's ability to produce accurate and reliable results while providing traceable justifications for its predictions, thus addressing key domain-specific knowledge and improving reliability in flavor development applications. Experimental results show that our proposed scientific agent method significantly outperforms baseline methods.

Our work represents a pilot study to design scientific agents for flavor profile sourcing tasks. We set a new benchmark for leveraging advanced AI technologies in the domain of flavor science. Our methodology provides clear, traceable insights into how flavor profiles are derived, thus facilitating greater acceptance and use of LLMs in practical food science applications. 

\section{Related Works}


\stitle{Retrieval-Augmented Generations} 
Retrieval-Augmented Language Models (RALMs) enhance the reasoning process by incorporating external knowledge sources, thus providing additional contextual information \cite{chen-etal-2017-reading, lee-etal-2019-latent, karpukhin-etal-2020-dense, pmlr-v162-borgeaud22a, zhang2024guidedprofilegenerationimproves}. Early works such as Retrieval-Augmented Generation (RAG; \citealt{NEURIPS2020_6b493230}) and Retrieval-Augmented Language Model pre-training (REALM; \citealt{pmlr-v119-guu20a}) have demonstrated significant improvements in contextual understanding and question-answering by fetching relevant documents from large corpora. Recent advancements in RALMs have significantly enhanced language models by integrating external knowledge dynamically and efficiently, improving performance with minimal retraining (\citealt{ram-etal-2023-context}; \citealt{jiang-etal-2023-active}; \citealt{10.5555/3648699.3648950}). In contrast to conventional RAG approaches retrieving from a static index base, our method leverages scholarly articles from the internet and augments model reasoning with consolidated information about flavor science. This approach innovates upon conventional RAG systems and highlights the extendability of our Flavor Scientific Agent.

\stitle{LLMs for Scientific Research} The application of large language models (LLMs) in scientific research has led to significant advancements across various disciplines. For example, domain-specific adaptations of BERT \cite{devlin-etal-2019-bert} for biology and biomedical domains have demonstrated significant performance improvements in tasks such as protein classification and text mining, underscoring the value of tailored pretraining for scientific corpora (\citealt{10.1093/bioinformatics/btz682}; \citealt{vig2021bertology}; \citealt{beltagy-etal-2019-scibert}; \citealt{10.1145/3458754}; \citealt{taylor2022galactica}). In the field of chemistry, LLMs have been tailored for tasks such as reaction prediction and molecule translation (\citealt{lu2022unified}; \citealt{edwards-etal-2022-translation}; \citealt{sagawa2023reactiont5}). However, they face common challenges, including a dependency on large, high-quality scientific datasets for training, which are often difficult to collect, and a lack of interpretability in their predictions, making it challenging for chemists to trust and use the results effectively (\citealt{horawalavithana-etal-2022-foundation}; \citealt{bran2023augmenting}). Our method leverages  retrieval augmented techniques to provide in-context learning at inference time, outperforming training-time methods. Our approach not only streamlines the process by generating and selecting the best hypothesis from multiple candidates but also enhances interpretability by providing explicit evidence of the rationales behind the outputs.

\section{Task and Data}

This section delineates the tasks and data designed to evaluate the efficacy of LLMs in predicting food items based on their molecular flavor profiles. These tasks are intended to assess the LLMs' capabilities in contextually-enhanced reasoning within the domain of molecular data.


\subsection{Task Definition}

\stitle{Molecular Food Prediction (MFP)} We define the task as learning the mapping function $h$ which takes a set of molecules $M = \{m_1, m_2, \dots, m_n\}$ as input and outputs a corresponding food source $F$. This can be expressed as:
\[
F = h(M) .
\]
This formulation aims to predict food sources based solely on their molecular composition, without reliance on specific algorithms or systems.

\stitle{Molecular Profile Completion (MPC)}  
The primary objective of the Molecular Profile Completion (MPC) task is to identify the missing molecules needed to complete the molecular profile of a given food item. The function $g$ receives the known food item $F$, a partial set of its molecules $M_{\text{partial}}$, and an integer $n$ representing the expected number of missing molecules. It then outputs the set of missing molecules $M_{\text{missing}}$. This can be mathematically represented as:
\[
M_{\text{missing}} = g(F, M_{\text{partial}}, n) .
\]
In this task, the integer $n$ is provided as part of the input, indicating the number of molecules that are missing. 
\subsection{Constructing the \textsc{FoodPuzzle} Dataset} 

\begin{figure}[t]
    \centering
    \small
    \includegraphics[width=0.5\textwidth]{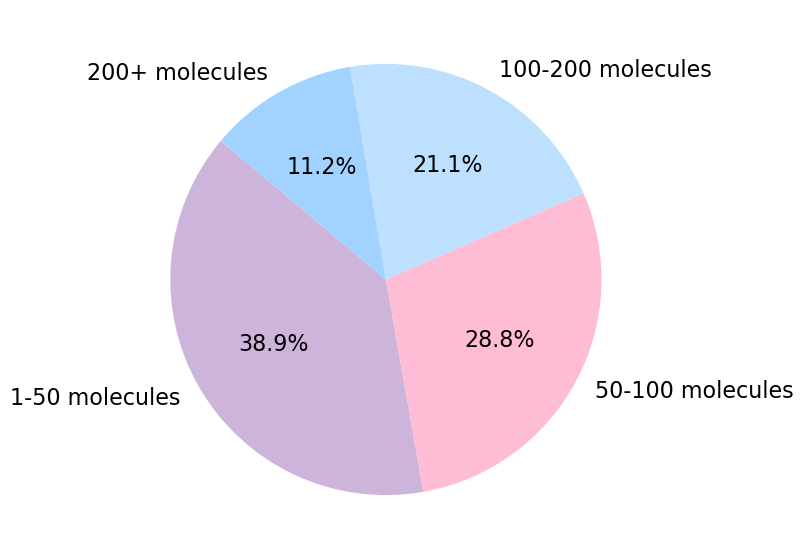}
    \caption{Distribution of the number of flavor molecules in the FOODPUZZLE dataset. }
    \label{fig:molecule_nums_statistics}
\end{figure}

\stitle{Data Collection}
To gather data for our study on flavor molecule prediction and analysis, we 
crawl through the extensive collection
of food entries of the the Flavor DB \footnote{The address of FlavorDB endpoint - \url{https://cosylab.iiitd.edu.in/flavordb}}. 
From the collected data, we create three distinct stores of information. The first store contains profile information pertaining to each molecule (e.g., 2-Ethylpyrazine), detailing its properties (e.g., isotope atom count)  and flavor characteristics (e.g., ordor). The second store focuses on information related to each food item, categorizing them by type and listing associated molecular compositions. Finally, we construct an association matrix that mapped which foods were associated with which molecules, represent as pairs of food and molecule identifiers. This structured approach ensures that our dataset was both comprehensive and well-organized, facilitating subsequent analysis and modeling efforts. The resulting \datasetname dataset maps 978 foods to 1766 flavor molecules. For the purpose of fine-tuning on the baseline, we split the dataset into train/dev/test sets by 80\%/10\%/10\%.




\begin{figure}[t]
    \small
    \centering
    \includegraphics[width=0.5\textwidth]{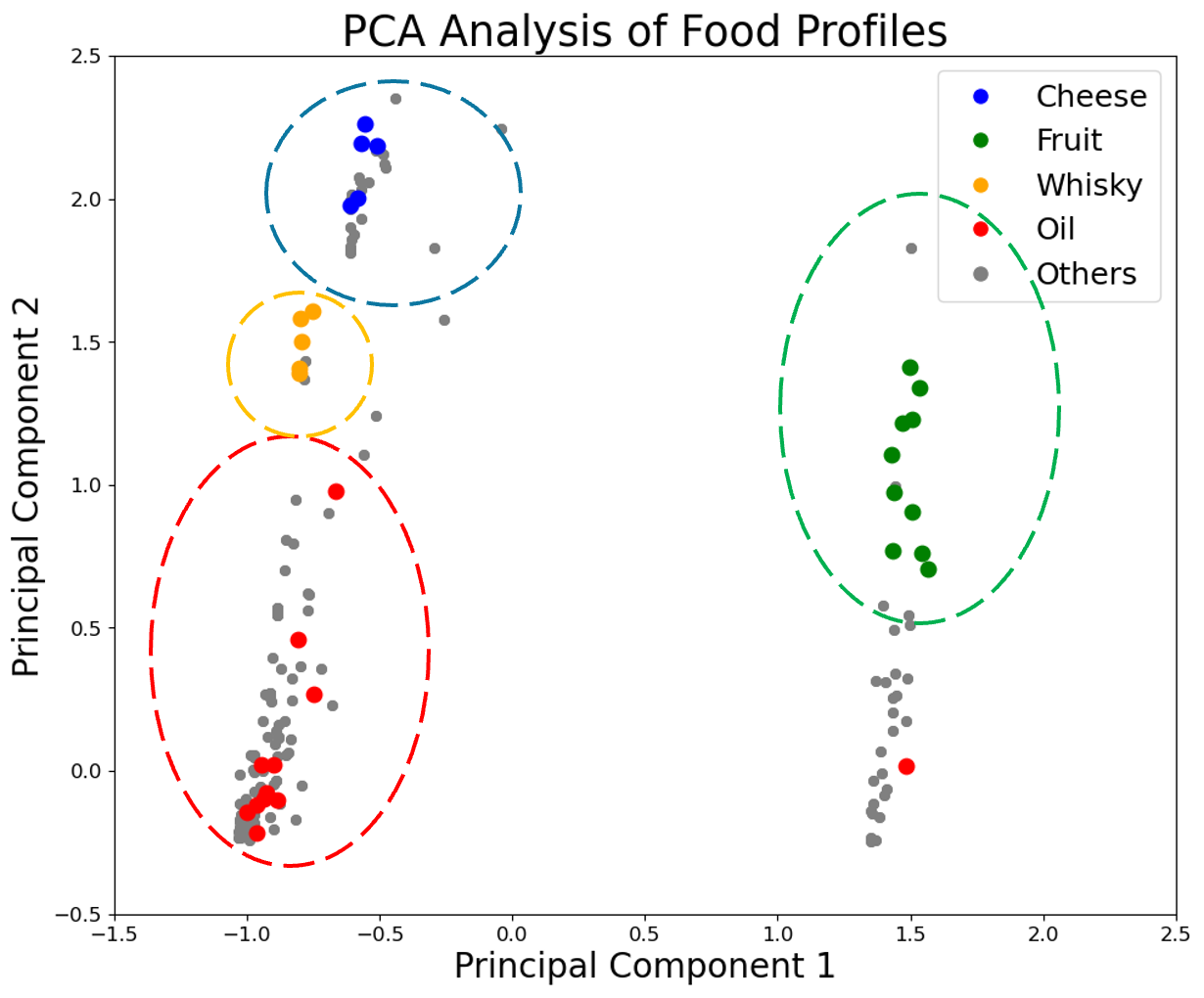}
    \caption{PCA visualization illustrating clustering of food entities based on molecular profiles.}
    \label{fig:pca_cluster}
\end{figure}


\stitle{Data Analysis}
Our dataset includes a variety of food entities, each associated with a unique set of molecules. 
Figure \ref{fig:molecule_nums_statistics} reveals that the majority of food entities are associated with a relatively small number of molecules, while a few entities have a significantly higher number of associated molecules. 
We employed feature encoding to represent each food item by its molecular composition, resulting in each item being characterized by a 1766-dimensional vector. Principal Component Analysis (PCA) of the encoded vectors, illustrated in Figure \ref{fig:pca_cluster}, reveals distinct clustering for categories such as cheese, whisky, oil, and fruit. These clusters indicate that food entities with analogous molecular compositions tend to aggregate, suggesting inherent patterns in flavor molecules that could potentially be predicted using machine learning techniques.

\subsection{Evaluation Protocols}
\stitle{Evaluation of Molecular Food Prediction (MFP)} 
The objective of MFP is to predict the category of a food item based on its molecular composition. There are 21 categories as shown in Table \ref{tab:food_categories}.  To measure the correctness of predicted food category, we provide the groundtruth answer $F^*$, which is sourced from FlavorDB. At evaluation time our models predict a food category $\hat{F}$ but this is done as a free text prediction; no ontology is provided. We thus compare  $\hat{F}$ and $F^*$ with a language model\footnote{In this work we use GPT-3.5-turbo; prompt details are provided in the Appendix B.} to assess whether the two are semantically equivalent. In this way we are able to match, e.g., \textit{``Alcohol''} and \textit{``Alcoholic Beverages''}, which are semantically equivalent but of different formats. We report accuracy as the metric for this task.


\begin{table}[h!]
\centering
\small
\begin{tabular}{|l l l|}
\hline
 & \textbf{Categories} & \\ \hline
Cereal             & Fruit               & Essential Oil       \\
Plant              & Bakery              & Fungus              \\
Seed               & Dish                & Spice               \\
Flower             & Nut and Seed        & Beverage            \\
Animal Product     & Vegetable           & Plant Derivative    \\
Additive           & Meat                &                     \\
Fish and Seafood   & Cereal Crop         &                     \\
Dairy              & Herb                &                     \\ \hline
\end{tabular}
\caption{Classification of food items into macro categories in the dataset.}
\label{tab:food_categories}
\end{table}

\stitle{Evaluation of Molecular Profile Completion (MPC)}  
MPC assesses the model's proficiency in predicting missing flavor molecules. 
To measure the correctness of predicted missing molecules, we provide the groundtruth answer set $M^*$. At evaluation time our models predict a set of molecules $\hat{M}$. We thus calculate the F1 score using $f(\hat{M})$ and $f(M^*)$, where $f$ extracts the set of functional groups of the provided set of molecules. Mathematically, the F1 score is calculated as below:
\[
F1 = \frac{2 \cdot |f(\hat{M}) \cap f(M^*)|}{|f(\hat{M})| + |f(M^*)|} .
\]

\noindent
By focusing on functional groups, this evaluation protocol leverages the fact that chemicals sharing functional groups tend to exhibit similar properties \cite{molecules21010001}. In this way we are able to prioritize chemical functionality over strict structural similarity, which ensures predictive relevance and streamline the evaluation process.

\begin{figure*}[t]
    \small
    \centering
    \includegraphics[width=0.9\textwidth]{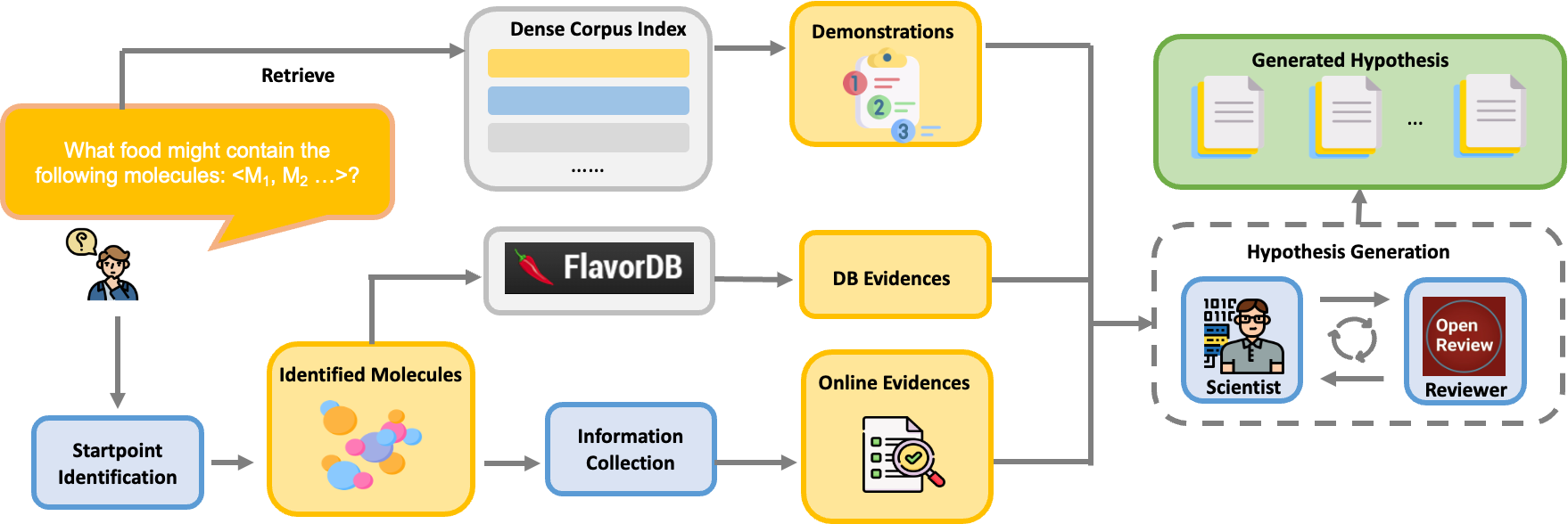}
    \caption{Architecture of the proposed Scientific Agent}
    \label{fig:architect}
\end{figure*}

\section{Evaluated Methods}

In this section, we explore selected baseline models ($\S$\ref{subsec:baselines}) and delve into the implementation details of our scientific agent designed for flavor development ($\S$\ref{subsec:our_approach}).

\subsection{Baseline Methods}
\label{subsec:baselines}
We evaluate the following baseline models to establish benchmarks for assessing the capabilities of our proposed scientific agent in the domain of flavor development. Our evaluation is organized into three distinct categories of models.

\stitle{Foundation LLMs} We evaluate ChatGPT-3.5-turbo\footnote{April 29th 2024 version.}, Gemini-1.5-Pro, and Llama-3-instruct-8B \cite{touvron2023llama} using zero-shot inference. Though being general-purpose and not specifically tailored to food science, these models are renowned for their robust performance across a broad spectrum of natural language processing tasks. The prompts used for zero-shot inference were carefully designed to ensure clarity and relevance to the tasks.


\stitle{Domain-specific LLMs} We recognize that molecule names and related chemical knowledge can be unfamiliar to general-purpose LLMs. Hence, we also consider models trained on corpora from related domains as baselines. These include MolT5, a model pretrained on a vast amount of unlabeled natural language text and molecule strings \cite{edwards-etal-2022-translation}, and BioT5, which is pretrained on structured and unstructured biology corpora 
for generative tasks involving molecular entities \cite{pei-etal-2023-biot5}. Since these two models are not instruction-tuned for conversational interaction, we follow \citet{wang-etal-2022-robust} to constrain the vocabulary set at decoding time and fine-tune these models on the training set for the task.

Both BioT5 and MolT5 are fine-tuned on our train set. We use 4 Nvidia A10G GPUs for our experiments. The models were fine-tuned with a batch size of 16 for the MFP task and 8 for the MPC task, using the AdamW optimizer with a learning rate of 0.00005, \(\beta_1 = 0.9\), and \(\beta_2 = 0.999\). Each model was trained for 20 epochs, with early stopping applied after 3 epochs based on validation loss.

\stitle{In-context Learning}
We also consider in-context learning techniques for our tasks. However, molecule names may be out-of-context information for LLMs due to limited exposure during pretraining, which can impair the performance of the agents. To enhance reasoning capabilities, we employ a retriever to selectively identify relevant data points from the labeled training set, providing supplementary information. For the implementation of this retriever, we evaluate several approaches: BM25 \cite{bm25}, Sentence Transformer \cite{reimers-2019-sentence-bert}, and Dense Passage Retrieval (DPR; \citealt{karpukhin-etal-2020-dense}). Our empirical results indicate that BM25 significantly outperforms dense retrieval methods, achieving a 23.2\% accuracy, compared to 13.1\% for the Sentence Transformer and 18.0\% for DPR in end-to-end performances on the MFC task. This superior performance of BM25 aligns with our expectations, given that dense retrievers are pre-trained and fine-tuned on generic natural language corpora, which may not effectively generalize to retrieve domain-specific data.


For both tasks, 
the BM25 retriever retrieves the top \(k\) most similar demonstrations to the query from a dense corpus index, where each passage is structured in the format: \textit{``Food: \(F'\). Molecules: \(M'\)''}. 
These retrieved demonstrations are then presented to the model as demonstrations, which are crucial for generating grounded and evidence-based output.


\subsection{Architecture of the Scientific Agent}
\label{subsec:our_approach}
We propose an advanced hybrid methodology that integrates Retrieval-Augmented Generation (RAG) with online scholarly sources and in-context learning, utilizing pertinent demonstrations. This approach is specifically tailored to surmount the challenges inherent in general-purpose LLMs that lack specialized knowledge in the domain of food science. The architecture of our scientific agent is designed to mirror the investigative processes typical of human scientists. This alignment not only enhances the agent’s ability to generate scientifically robust hypotheses but also improves the explicability of errors, thereby contributing to its operational transparency and efficacy in flavor science tasks.


\stitle{Starting Point  Identification} Similarly to solving other puzzles, it is important to identify suitable starting points in our flavor profile sourcing task. Consequently, for the MFP, given a list of flavor molecules, denoted as $M$, we focus on selecting molecules that are pivotal for our analysis. We calculate the information entropy for each molecule in $M$ based on the frequency of appearances in the train set and select up to 10 molecules with the lowest information entropy to identify an informative subset of molecules. These selected molecules $M^*$ then serve as starting points for subsequent investigations.


\begin{table*}[ht!]
\centering
\begin{tabular}{ccccc}
\toprule
\small
\textbf{Category} & \textbf{Model} & \textbf{MFP accuracy(\%)} & \textbf{MPC F1 score} \\
\midrule
\multirow{2}{*}{Domain-Specific LLMs} & MolT5 \cite{edwards-etal-2022-translation} & 9.8 & 0.144 \\
                                      & Bio-T5 \cite{pei-etal-2023-biot5} & 16.6 & 0.278 \\

\midrule
\multirow{3}{*}{Zero-shot} & LLaMA3-8B-instruct & 15.0 & 0.292 \\
                                 & Gemini1.5 Pro & 19.0 & 0.340 \\
                                 & GPT-3.5 Turbo & 12.2 & 0.327 \\
\midrule
\multirow{3}{*}{In-context Learning} & LLaMA3-8B-instruct & 31.6 & 0.349 \\
                               & Gemini1.5 Pro & 34.6 & 0.373 \\
                               & GPT-3.5 Turbo & 23.2 & 0.360 \\


\midrule
\midrule
\multirow{1}{*}{Scientist Agent} & LLaMA3-8B-instruct & \textbf{35.5} & \textbf{0.374} \\
                               & Gemini1.5 Pro & 34.2 & 0.333 \\
                               & GPT-3.5 Turbo & 26.9 & \textbf{0.374}\\
\bottomrule
\end{tabular}
\caption{Combined Results for Model Accuracies in Molecule Food Prediction and F1 Scores for Molecule Profile Completion}
\label{tab:combined_results}
\end{table*}

\stitle{Information Collection} One key challenge in employing general-purpose LLMs in specialized fields like food science is their inherent limitation in domain specificity. While these models are trained on massive corpora and varied task datasets that provide a wide range of general knowledge, they often lack the detailed, specific insights needed to tackle complex, domain-focused issues effectively. This shortfall is particularly noticeable in food science, where the models struggle to accurately comprehend or predict the subtle nuances of flavor profiles and the interactions among different compounds. Though in-context learning introduces relevant demonstrations to the agent, it still predominantly relies on the parametric knowledge of LLMs to figure out the relationship between demonstrations and test instances. This inherent limitation can result in outputs that may not sufficiently use in-context information.

To address this gap, Out-of-Domain Evidence is sourced from scholarly articles and internet blogs in addition to the in-context demonstrations. Specifically, for the MFP task, the process begins by iteratively querying each molecule \(m \in M^*\).
The query is formulated as: \textit{``What are the common food sources that could contain \(m\)?''} Similarly, for the MPC task, the process begins by querying each food item \(F\), with the corresponding query: \textit{``What molecules are present in \(F\)?''} We then use the Google Custom Search API to identify academic articles that detail which plants, animals, or other organisms naturally produce these molecules. To ensure the reliability of the information retrieved, the search is confined to scientific repositories such as PubMed \footnote{https://pubmed.ncbi.nlm.nih.gov} and arXiv \footnote{https://arxiv.org}. The scientific agent gathers the sourced articles, storing the information as evidence. To enhance efficiency during inference, the Google Custom Search is executed offline, and the results are stored locally. Consequently, during inference, the system simply retrieves the relevant information from local storage. 

\stitle{Hypothesis Generation} The final phase is hypothesis generation. We realize that it is possible for the collected evidences to sometimes contradict each other, pointing to different answers. Leveraging the Chain-of-Thought (CoT) approach \cite{NEURIPS2022_9d560961}, we adopt a role-play based agent framework to process sourced evidences and generate the final hypothesis. Specifically, we initialize two role-based models: the Scientist that proposes three hypotheses and the Reviewer that evaluates the argument quality and selects the best hypothesis
\footnote{We include prompt examples in Appendix B}. 
In the MFP task, we retrieve locally stored evidence from FlavorDB related to up to ten key molecules identified during the initial point identification phase. This evidence, along with relevant data points, is then provided to the Scientist model. For the MPC task, as there is no starting point identification process, we retrieve locally evidence related to the food item directly from FlavorDB, and this evidence is provided to the Scientist model. Also, as implemented in the in-context learning baseline, a BM25 retriever is used to select the three most similar demonstrations, which are also given to the Scientist model. The Scientist, therefore, receives data input from \datasetname, the retrieved evidence, and demonstrations, and uses these to generate three hypotheses. The Reviewer model then receives the data input, samples, and the three hypotheses generated by the Scientist. After reviewing the hypotheses, the Reviewer might reject or select the most suitable one. Finally, the scientific model returns this best hypothesis as the final prediction. 

\section{Experiments}

In this section, we report and analyze the benchmarking results ($\S$\ref{subsec:main_reults}). We randomly select 50 prediction errors and manually inspect them. We present a comprehensive error
analysis that outlines key categories of errors as insights for future model refinement ($\S$\ref{subsec:error_analysis}).

\subsection{Main Results}
\label{subsec:main_reults}
Table \ref{tab:combined_results} presents the combined results for model accuracies in Molecule Food Prediction (MFP) and F1 scores for Molecule Profile Completion (MPC). 
Domain-Specific LLMs such as MolT5 and Bio-T5 exhibit lower performance in both MFP accuracy and MPC F1 score compared to other models. The limitation of these models can be attributed to two reasons. On the one hand, their specialization in molecular and biological text does not necessarily translate to broader applications in food science. However, small parameter sizes and lack of instruction tuning might also limit model performance.

In the zero-shot scenario, foundation LLMs like LLAMA3-8B-Instruct, Gemini1.5 Pro, and GPT-3.5 Turbo show varying degrees of effectiveness. LLAMA3-8B-Instruct achieves an MFP accuracy of 15.0\% and an F1 score of 0.292, indicating moderate performance. Gemini1.5 Pro performs better with an MFP accuracy of 19.0\% and an F1 score of 0.340, demonstrating its relative strength in both food prediction and molecule profile completion. GPT-3.5 Turbo balances its performance with an MFP accuracy of 12.2\% and an F1 score of 0.327. These results highlight the challenging nature of our tasks and suggest that while general-purpose models can achieve reasonable performance, they still lack the domain-specific knowledge required for food science tasks.

Leveraging in-context learning techniques, LLMs demonstrate notable improvements over the foundation and domain-specific LLMs. 
The enhanced performance of these models underscores the importance of in-context learning in providing relevant additional information and improving prediction accuracy. Furthermore, these results highlight the effectiveness of our \datasetname dataset in enabling models to take advantage of the relevant context for better predictions.

The Autonomous Scientist Agent exhibits superior performance in both Molecular Food Prediction accuracy and F1 scores for Molecular Profile Completion, compared to other model categories. The LLaMA3-8b-instruct model within the agent achieves an MFP accuracy of 35.5\%, outperforming the best in-context learning model (Gemini1.5 Pro with 34.6\%) and the highest foundation LLM (Gemini1.5 Pro with 19.0\%). Furthermore, the agent shows robust F1 scores, with LLaMA3-8b-instruct at 0.374, Gemini1.5 Pro at 0.333, and GPT-3.5 Turbo at 0.374, indicating its balanced strength in both tasks. 
This approach effectively addresses the challenges faced by general-purpose LLMs that lack domain-specific knowledge in food science. By sourcing and consolidating domain-specific evidence and employing a structured approach to hypothesis generation, the agent enhances prediction accuracy and molecule profile completion.

\subsection{Error Analysis}
\label{subsec:error_analysis}
Next, we define a taxonomy of error types discerned among evaluated scientific agents, based on an analysis of fifty randomly selected prediction errors. It is possible that a single example may exhibit multiple error types concurrently. To preclude redundancy, only the most salient error per example is accounted for in this analysis.

\stitle{Inappropriate Initialization of Search Space (32\%)}
When identifying key molecules to understand flavor profiles, the autonomous scientific agent often struggle due to limited domain-specific knowledge. For example, while domain experts can focus on molecules such as ``hydrogen sulfide'' and ``diethyl sulfide'' for their significant roles in foods like eggs, LLMs tend to suggest more generic molecules such as thiamine and betaine. These are common across various foods and can lead the agent to inaccurate, fruit-biased interpretations by narrowing the scope of analysis. This mismatch highlights the importance of incorporating specific and expert-driven insights into the training and inference of LLMs used in flavor science.

\stitle{Epistemic Hallucination (26\%)}
The phenomenon of hallucination was observed across the evaluated LLMs. During processes intended to enhance model reasoning through chain-of-thought techniques, the output generated by the agent often lacks grounding in factual evidence. For example, when asked to identify flavor molecules for eggs, an LLM might list molecules commonly found in eggs rather than those that could recreate the egg flavor profile. This distinction is crucial: The goal is to identify molecules that can mimic or replicate flavors, not just those that occur naturally in the food item. Effective prompting can mitigate this issue. Clearly specifying that the task involves identifying molecules capable of replicating specific flavors, rather than listing inherent molecules, can direct the LLMs to focus correctly. Additionally, providing explicit examples and context in the prompts can further align the LLM responses with the intended task.

\stitle{Wrong Interpretation of Online Sources (20\%)}
Another considerable source arises from the misinterpretation of scholarly articles during the process of sourcing evidence online. When models asked to perform aspect-summarization toward input queries, they exhibits a tendency to ``force'' answers where direct evidence might be lacking or ambiguous. This compulsion to generate conclusions can lead to erroneous or overly speculative assertions. For instance, consider a scenario where a scholarly article states that ethanethiol is characterized by more roasted and toasted notes, and may exhibit a coffee-like character. LLMs, in its attempt to generate actionable insights, might erroneously infer and assert that ``coffee contains Ethanethiol as one of its flavor molecules.'' Such a statement is a speculative leap rather than a factually supported conclusion, reflecting the model's predisposition towards generating responses even in the absence of clear evidence.

\section{Future Opportunities}

Integrating autonomous flavor scientists and \datasetname into the R\&D pipeline offers significant advancements in flavor science, sensory science, and food product development. In this section we discuss how collaboration with wet lab and sensory scientists to evaluate chemical, biological, and sensory properties of flavor compounds will enhance model accuracy and impact. 

\stitle{Chemical Synthesis and Analysis} 
Using analytical instruments like LC-MS and GC-MS to detect and quantify flavor molecules in food samples allows for verification of structure and purity, validating flavor profile hypotheses. Synthetic chemistry labs can synthesize predicted candidate flavor molecules, enabling verification of their organoleptic properties. Sensory labs can manage human sensory panels to evaluate synthesized flavors, fine-tuning AI model predictions to align with human perceptions of flavor quality and intensity.

\stitle{Biological Analysis} 
Biological laboratories can perform bioassays and in vitro testing to assess the safety and efficacy of flavor compounds. High-throughput screening (HTS) techniques can rapidly test large libraries of flavor molecules, generating extensive datasets to enhance AI predictions.

\section{Conclusion}
This paper underscores the transformative potential of LLM agents in flavor science, particularly through the development of an scientific agent tailored for flavor profile sourcing. Key contributions include the formulation of flavor profile sourcing and understanding as a LLM agents task, the creation of the \datasetname dataset from FlavorDB, and the creation of a comprehensive scientific agent pipeline that can perform the labor-intensive tasks of flavor sourcing and understanding with enhanced efficiency and precision. Our findings demonstrate that our method outperforms traditional models and provides traceable and reliable predictions. This work not only sets a new benchmark for the application of AI in flavor science but also paves the way for further technological advancements.



\FloatBarrier

\bibliography{aaai25, custom}

\clearpage

\end{document}